# A HIGH QUALITY AND PHONETIC BALANCED SPEECH CORPUS FOR VIETNAMESE

*Pham Ngoc Phuong[1], Quoc Truong Do[2], Luong Chi Mai[3]*

[1]Information Technology Center, Thai Nguyen University; [2]Nara Institute of Science and Technology (NAIST);
[3]Institute of Information Technology, Vietnamese Academy of Science and Technology
[1]phuongpn@tnu.edu.vn; [2]do.truong.dj3@is.naist.jp; [3]lcmai@ioit.ac.vn

**ABSTRACT**

This paper presents a high quality Vietnamese speech corpus that can be used for analyzing Vietnamese speech characteristic as well as building speech synthesis models. The corpus consists of 5400 clean-speech utterances spoken by 12 speakers including 6 males and 6 females. The corpus is designed with phonetic balanced in mind so that it can be used for speech synthesis, especially, speech adaptation approaches. Specifically, all speakers utter a common dataset contains 250 phonetic balanced sentences. To increase the variety of speech context, each speaker also utters another 200 non-shared, phonetic-balanced sentences. The speakers are selected to cover a wide range of age and come from different regions of the North of Vietnam. The audios are recorded in a soundproof studio room, they are sampling at 48 kHz, 16 bits PCM, mono channel.

## 1. INTRODUCTION

In speech related research, especially, in speech analysis and speech synthesis studies, it is crucial to have high quality speech corpora. In some popular languages such as English and Japanese, there have been many intensive researchs on designing databases [1,2]. In Vietnam, the number of speech data research is alo increasing. There are many speech corpora such as VOV (Radio broadcast resources) [3], MICA VNSpeechCorpus [4], AIlab VIVOS [5], VAIS-1000 [6]. However, those corpora are either small or do not have a high quality sound. In particular, the VAIS-1000 corpus is designed from only a speaker with local accent from one particular region; The VIVOS corpus does not have high quality speech and it is designed specifictly for speech recognition tasks. The audio from VOV corpus are selected from media sources and are also only suitable for speech recognition tasks.

On the other hand, while research on speech synthesis adaptation which we can generate a model for a specific speaker with a very limited amount of speech samples is an active research field on popular languages [7], it is difficult to conduct such a research on Vietnamese due to a very high requirement on the data design. First, audio has to be recorded in a clean, soundproof recoding room to ensure the high quality speech. Second, the speakers have to be selected to cover wide range of ages as well as living areas. The corpus proposed in MICA VNSpeechCorpus [4] is well designed and contains good quality speech. However, although the total size of the corpus is big, the amount of short sentences that are suitable for speech synthesis adaptation is rather small.

In this paper, we present a high quality and large scale Vietnamese speech corpus. We design the corpus with a strategy that maximizes the coverage of monophone and biphone. Speakers are carefully selected with a wide range of age and living region. In the following section, we first describe Vietnamese phonetic structure (Section 2), it provides essential information to design and select the recording transcription described in Section 3. Finally, we provide details of the corpus along with analyses in Section 4.

## 2. BASIC PHONETIC STRUCTURE OF VIETNAMESE

Vietnamese language is a complex language compared with other languages because it is a monosyllable language with tones, every syllable always carries a certain tone [8,9]:

| Initial | TONE | | |
|---|---|---|---|
| | FINAL | | |
| | Onset | Nucleus | Coda |

**Table 1.** Structure of Vietnamese syllables

There are 22 initials in Vietnamese, include: /b, m, f, v, t, t', d, n, z, ʐ, s, ʂ, c, ʈ, ɲ, l, k, χ, ŋ, ɣ, h, ʔ/. Onset /w/ has a function of lowering the tone of the syllables. The number of main finals consists of 16 phonemes, including 13 vowels and 3 dipthongs. Specifically, /i, e, ɛ, ɤ, ɤ̆, a, ɯ, ă, u, o, ɔ, ɔ̆, ɛ̆/ and 3 dipthongs / ie, ɯɤ, uo/. In addition to the final /zero/, there are 8 positive finals including 6 consonants /m, n, ŋ, p, t, k/ and 2 semi-vowel /-w, -j/.

There are 6 tones in Vietnamese. Five tones are represented by different diacritical marks such as l*ow*- falling tone, high-broken tone, low-rising tone, high-rising tone, low-broken tone. The tone called mid tone is not represented by a mark. Tones are differentiated in the following Table 2 [10,9]:

| Contour Pitch | Flat | Unflat | |
|---|---|---|---|
| | | Broken | Unbroken |
| High | No mark | High-broken | High-rising |
| Low | Low-falling | Low-rising | Low-broken |

**Table 2.** Structure of Vietnamese tones

The total number of unique syllables in Vietnamese is 19000 but there are only 6500 syllables used in practice [8,9].

## 3. DESIGN

In this section, we describe our recording transcription design strategy. To have a wide variety of context, we select the recording sentences from electronic newspapers. However, since the data from newspapers is typically noisy, we need to put it through a chain of data processing phases as illustrated in Figure 1. Then, we select the smallest subset of data that maintain the phonetic balance criterion.

### 3.1. Text data preprocessing

The text recording needs to be designed to meet the criteria that it is not too large but it must ensure the phonetic balance. We first collect a large amount of text from electronic newspapers. And then, process it in a chain of processing phases.

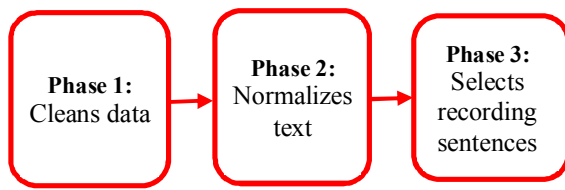

*Figure 1. The process of refining and processing the recording texts*

**Phase 1:** The data downloaded from newspapers has the main content stored in the <content> tags, other information outside the tag is metadata. To reduce noises, we only select texts from the <content> tag. Next we cut them into small sentences based on ending punctuations such as ".", "?", "!". To further reduce noises, we remove all lines contain post time, shortcuts, address, as well as arbitrary strings (asterisks, special characters, punctuation marks, author names, annotations, quoted source names etc).

**Phase 2**: In this phase, the main task is to normalize the text (includes many non standard words) according to the standards words in Vietnamese [11]. Non standard words include digit sequences; numbers; abbreviations; units of measurement; roman numerals; foreign proper names and place names... We analyzed text and used the technique to transforming (or expanding) a sequence of words into a common orthographic transcription. The process is done by 2 steps:

*Step 1.* To reduce pronunciation ambiguity, all numbers, date, time and measure units are spelled out with the following rules:

- Number format:

Numbers are transcribed in code by assigning them to arrays and transcribing them into corresponding strings (e.g. 1235 → một nghìn hai trăm ba mươi năm). Then exceptions are replaced with standard words (e.g. không mươi → lẻ, mươi năm → mươi lăm, mươi một → mươi mốt).

- Time format:

Format dd/mm/yyyy is automatically transcribed into day...month...year .

Format (dd/mm, dd-mm-yyyy and dd-mm) with the word 'day' standing in front is transcribed as "day", "month".

Format hh:mm:ss is understood as hour, minute, second

Format hh:mm with "at" standing in front is transcribed into "hour", "minute".

- Units of measurement:

Separate alphanumeric characters with spaces (e.g. 10Kg → 10 k, 10m → 10 m, 11hz → 11 hz, 8/10 → 8 / 10, 90% → 90 %).

Then, replace words with transcribing digits for signs or measure units (e.g. 10 kg → ten kilograms, 10 meters → ten meters, 11 hz → eleventh hertz, 8 / 10 → eight per ten, 90% → ninety percent).

*Step 2.* Transcribe abbreviated acronyms or proper names with self-defined dictionaries (e.g. TP → city, HCM → Ho Chi Minh City, VND → Vietnam dong, Paris → Pa ri, Samsung → Sam Sung).

After normalizing the text, we split them into small sentences and only keep ones contain minimum 40 and maximum 90 syllables. This is an appropriate length for speech recording.

**Phase 3:** The final step is to select a good amount of sentences for audio recording. The recording sentences should maintain the phonetic balance property and be small to reduce recording cost. We adopt text selection based on greedy search to find the optimal sentences [12]. This step is repeated until a certain amount of sentences are selected.

### 3.2. Recording

To help speeding up the recording, as well as make it easier and less prone to human error. We designed a recording application as shown in Figure 2. The speaker can listen to their recorded audio and can also see the audio signal to ensure that there is 1 second of silence at the beginning and ending of utterances and there is no audio clipping occurred. During a recording session, if there is any sentence doesn't meet the requirement, the speaker will only need to record the sentence again. The quality assurance process is managed by an administrator using the same application with speakers.

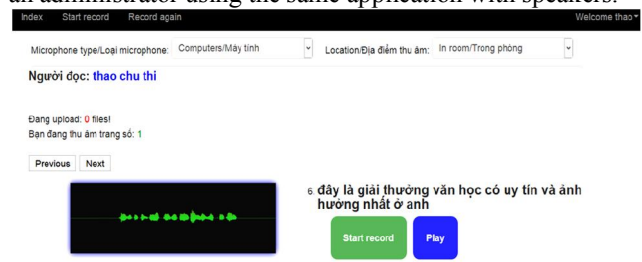

*Figure 2. Web-based recording application*

The audio is recorded with a high-quality TakStar PC-K600 microphone using a Windows 10 computer running a Firefox browser. The audio sampling rate is 48Hhz, 16-bits PCM, and mono channel.

## 4. EXPERIMENT

The following Table 3 summarizes the data collected in the above text data preprocessing section:

| No | Name | Size | Unit | Note |
|---|---|---|---|---|
| 1 | Original data | 1.978/9,6 | File/GB | Data in file format .txt |
| 2 | Phase 1 | 1.978/8,1 | File/GB | Refine content, raw processing |
| 3 | Phase 2 | 4.036.312 | Sentence | Text Normalization |
| 4 | Phase 3 | 250/2.400 | share/non-shared sentences | Select sentences for recording: each person records 250 share common sentence and 200 none-share sentences. |

**Table 3**. Steps of text data preprocessing

### 4.1 Corpus detail

To ensure the wide variety of speech, we selected 12 speakers from 5 provinces of North Vietnam including 6 males and 6 females aged 22 to 35. Our target number recording sentences is 250 that has to maintain the phonetic-balanced property. Our experiment on data selection algorithm described in the section above shown that only 250 utterances are needed to cover all monophone and 99% of bi-phone. This is a good sign that we do not have to select many sentences to meet the requirement. We use this 250 utterances for all speakers.

However, to increase the variety, we also want each speaker utter a separate set of text. Therefore, we run the data selection algorithm again to select other sets for each speaker. Each run, we removed the selected sentence to make sure there are no duplicated sentences in the corpus. As the result, we have 200 x 12 phonetic-balanced sentences. Note that male and females speakers share the same recording utterances. As the result, we have recorded 2,400 speech samples. The results as shown in Table 4.

| Data set | Number of letters | Number of syllables | Number of syllables per sentence | Unique syllables |
|---|---|---|---|---|
| 250 sentence set | 14.954 | 3.268 | 59,8160 | 1.205 |
| 2400 sentence set | 120.6320 | 26.308 | 50,2633 | 2.758 |

**Table 4**. Statistics of text sentences for recording

### 4.2 Data analysis
#### 4.2.1 Phonemes statistic

To evaluate data corpus, we use several modules to count two text data sets based on occurrence frequency and deference of phonemes, syllables and words. The results as shown in Table 5.

| | Set of 250 share common sentences | | | | Set of 2,400 none-share sentences | | | |
|---|---|---|---|---|---|---|---|---|
| No | Bi phone | Frequency of Occurence | Mono phone | Frequency of Occurence | Bi phone | Frequency of Occurence | Mono phone | Frequency of Occurence |
| 1 | ea-ngz | 89 | ngz | 526 | a-iz | 809 | a | 4482 |
| 2 | a-iz | 84 | a | 510 | oo-ngz | 644 | ngz | 4235 |
| 3 | oo-ngz | 78 | iz | 347 | ea-ngz | 636 | iz | 3133 |
| 4 | l-a | 76 | nz | 340 | aa-nz | 507 | nz | 2871 |
| 5 | oa-ngz | 59 | k | 296 | ie-nz | 504 | k | 2432 |
| 6 | aa-nz | 58 | i | 286 | u-ngz | 484 | oo | 2355 |
| 7 | ngz-k | 56 | oo | 265 | l-a | 482 | i | 2286 |
| 8 | u-ngz | 54 | dd | 236 | a-nz | 472 | dd | 1981 |
| 9 | k-o | 53 | tr | 232 | aw-iz | 469 | tr | 1863 |
| 10 | aw-iz | 52 | aa | 227 | oo-iz | 464 | aa | 1824 |
| 11 | a-nz | 51 | wa | 218 | i-ngz | 447 | kc | 1724 |
| 12 | ie-uz | 51 | kc | 216 | w-a | 436 | aw | 1658 |
| 13 | wa-ngz | 50 | ie | 211 | oa-ngz | 419 | ie | 1647 |
| 14 | aa-tc | 49 | aw | 202 | k-o | 411 | wa | 1598 |
| 15 | ow-iz | 49 | ee | 190 | ie-uz | 401 | uz | 1579 |
| 16 | ie-nz | 48 | uz | 188 | ngz-k | 400 | o | 1464 |
| 17 | uw-ngz | 48 | o | 183 | k-uo | 389 | t | 1443 |
| 18 | w-a | 45 | uw | 182 | ow-iz | 389 | mz | 1442 |
| 19 | wa-kc | 45 | th | 180 | b-a | 386 | ee | 1410 |
| 20 | oo-iz | 44 | tc | 175 | uw-ngz | 379 | m | 1386 |

**Table 5**. Statistics of 20 most popular phonemes in 2 data set (without sil)

#### 4.2.2 Sound quality analysis

The sound quality was analyzed by Praat v6.0 software to evaluate the characteristics of sound waves, spectra, pitch and sound intensity [13]. The following example in Figure 3 will analyze the waveforms and spectrograms of a female voice extracted from the utterance "đây là một chuyên mục tôi cảm thấy tâm đắc và uy tín" (in English "this is the prestige category which I feel favorite").

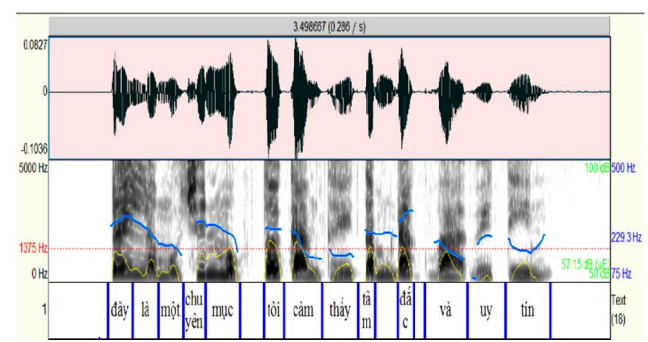

*Figure 3. Waveform and spectrogram of the female voice*

Data evaluation was done through the assessment of the recording environment, the noise ratio [14]. Through analysis and evaluation of all data, the recording was evaluated to be of good quality with clear sound and little noise.

#### 4.2.3 Duration analysis

In this experiment, we are interested in the difference between genders and ages in term of duration of words. To obtain word duration, we build an automatic speech recognition (ASR) using Kaldi toolkit [15]. The training for the ASR system is the same data used for the decoding process so that we can have accurate audio alignment results. State duration of each HMM are modeled by a multivariate Gaussian estimated from histograms of state durations which were obtained by the Viterbi segmentation of training data [16,17].

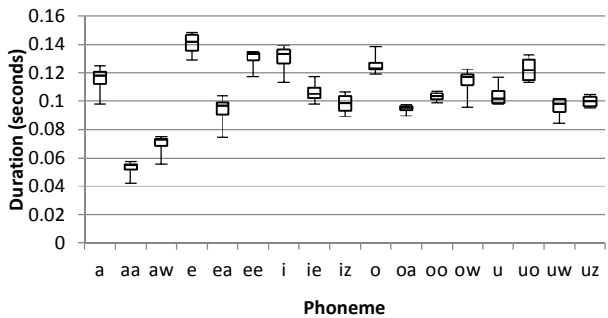

*Figure 4. Duration distributions of vowels spoken by female voices at the same age*

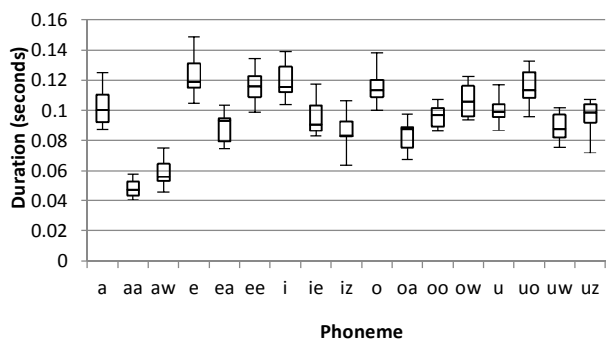

*Figure 5. Duration distribution of vowels spoken by people with wide range of age and different gender*

Figure 4 shows vowel duration distributions of 4 female speakers at the same age. As we can see, the range of all distributions are quite small, indicating that females at the same age tend to have similar reading speed. On the other hand, Figure 5 shows the duration distributions of speakers with different genders and wide range of age. We can clearly see that the distribution is larger than it is in Figure 4.

As one of the purpose of the corpus is to build speech synthesis adaptation systems. The result indicates an important clue that we should not use as many data as possible but instead only use data that have similar characteristic such as gender or age to achieve optimize results in term of duration adaptation.

## 5. CONCLUSION

In this paper, we have described a high quality speech corpus for Vietnamese that is suitable for data analysis and constructing speech synthesis systems. The work result is a high-quality data set that contains 5,400 utterances with the accompanied text which were recorded by variety gender and age speakers. This is the minimum data set that meets our target which guarantee that the amount of short sentences with phonetic balanced is suitable for speech synthesis adaptation. Future work will focus on expanding the data size in both term of speakers and accent, as well as, utilizing the corpus to construct Vietnamese speech synthesis adaptation systems.